\documentclass[runningheads]{llncs}
\usepackage{graphicx}
\usepackage{tabularx} 
\usepackage{booktabs}
\usepackage{multirow}
\usepackage{multicol}
\usepackage{todonotes}
\begin{document}
\title{Generating Topic Pages for Scientific Concepts Using Scientific Publications}
%
%

\author{%
Hosein Azarbonyad  \and 
Zubair Afzal  \and 
George Tsatsaronis
}%

\institute{ 
Elsevier, The Netherlands\\
\email{\{h.azarbonyad, zubair.afzal, g.tsatsaronis\}@elsevier.com}
}

\maketitle              
\begin{abstract}
In this paper, we describe \emph{Topic Pages}, an inventory of scientific concepts and information around them extracted from a large collection of scientific books and journals. The main aim of \emph{Topic Pages} is to provide all the necessary information to the readers to understand scientific concepts they come across while reading scholarly content in any scientific domain. \emph{Topic Pages} are a collection of automatically generated information pages using NLP and ML, each corresponding to a scientific concept. Each page contains three pieces of information: a definition, related concepts, and the most relevant snippets, all extracted from scientific peer-reviewed publications. In this paper, we discuss the details of different components to extract each of these elements. The collection of pages in production contains over $360,000$ \emph{Topic Pages} across $20$ different scientific domains with an average of $23$ million unique visits per month, constituting it a popular source for scientific information.

\end{abstract}

\section{Introduction}
Technical terminology is an important piece of scientific publications \cite{jin2013mining,kang2020document}. 
Scientists and researchers use technical terminology and concepts to convey information concisely. As a result, there is an overwhelming and growing number of scientific concepts in any scientific domain, adding to the difficulties scientists have to catch up with the ever-growing list of technical concepts and new content.
Knowledge sources such as \emph{Wikipedia} can provide useful information on technical and scientific concepts to a large extent, however, due to their \emph{``wisdom-of-crowds"} creation method there are many omissions and errors, and they may not always be a trustworthy source to understand and refer to a scientific concept. Our \emph{Topic Pages}\footnote{\url{https://www.elsevier.com/solutions/sciencedirect/topics}} proposition creates a knowledge source in a \emph{``wisdom-of-experts"} fashion, as the information on scientific concepts is extracted from iconic scientific books in the domain, or from high-impact peer-reviewed scientific publications on the topic.

\begin{figure}[t]
\centering
\includegraphics[width=10cm, height=5cm]{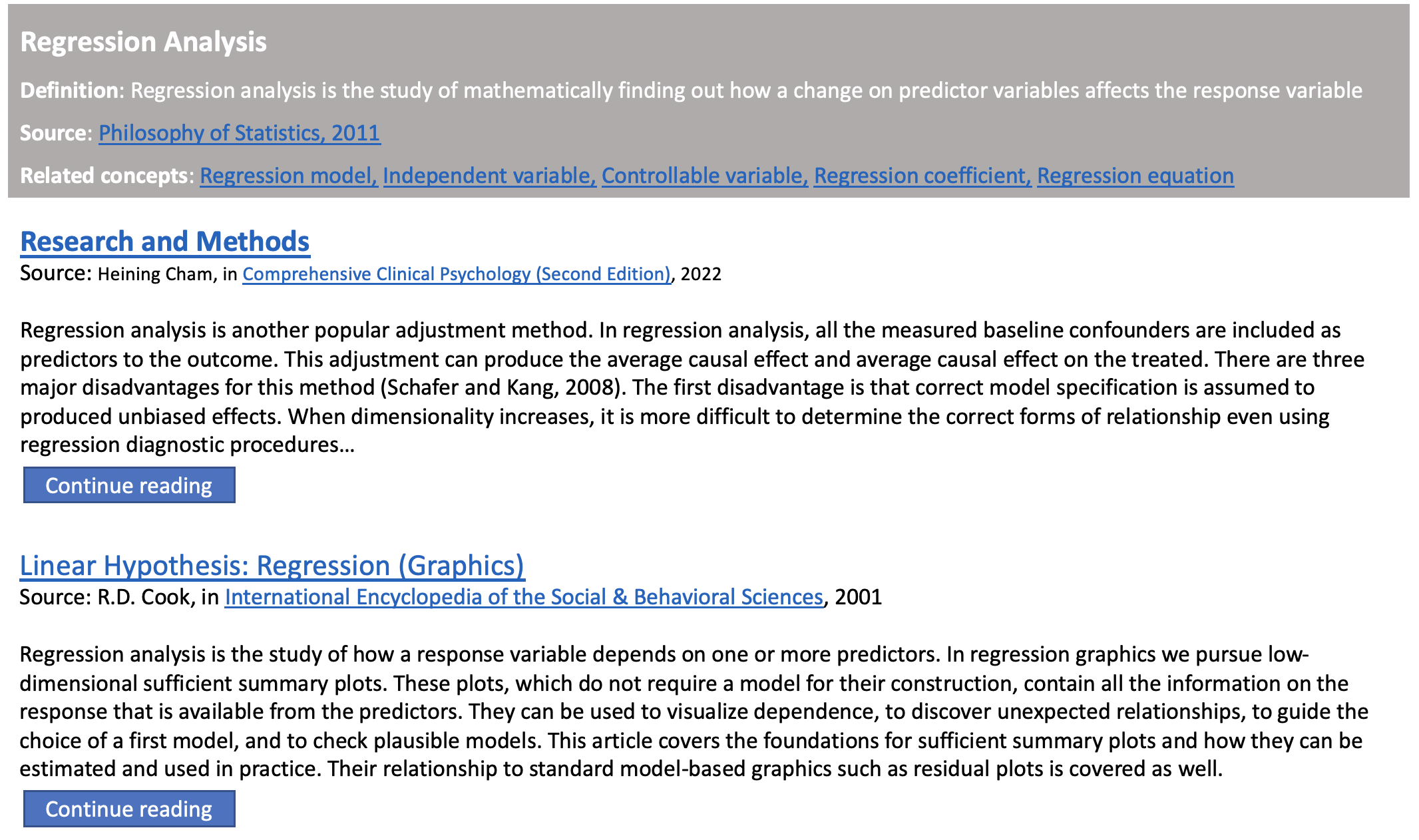}
\vspace{-10pt}
\footnotesize
\caption{An example \emph{Topic Page} presenting the concept ``regression analysis''', with a definition, related concepts, and a set of relevant snippets extracted from articles and books.}
\label{fig:example}
\end{figure}


Each \emph{Topic Page} is centered around one scientific concept and contains a definition for the concept, a set of related concepts, and a set of relevant snippets all extracted from scientific peer-reviewed articles and books. The definition comprises one sentence extracted from books and journals that provides a brief, yet concise, description of the concept. 
Snippets are text excerpts from books or journals, relevant to the concept, and provide contextual information about the concept. 
Related concepts are a set of most relevant concepts to the given concept that can help users to explore the relevant terminology around their concept of interest. 

The collection contains over $360,000$ \emph{Topic Pages} in $20$ different scientific domains. These topic pages are hyperlinked from publications in ScienceDirect\footnote{\url{https://www.sciencedirect.com/}}, which is one of the largest scientific publication search engines and databases containing over $18$ million full-text articles, helping users to navigate to the corresponding \emph{Topic Page} when they encounter an unfamiliar scientific concept in an article with just one click. There are over $5.8$ million articles that provide hyperlinks that we have created from scientific articles to topic pages. \emph{Topic Pages} attract over $23$ million unique visits per month. 

In the remainder of the paper, we briefly review related work in Section \ref{sec:rel-work}, we describe the technical pipeline for generating \emph{Topic Pages} in Section \ref{sec:pipeline}, we evaluate empirically the most challenging module of the pipeline, which is the definition extraction, in Section \ref{sec:results} and we conclude in Section \ref{sec:limits} by arraying some limitations of the current technical solution and provide pointers to future work.   

\section{Related Work}
\label{sec:rel-work}
To the best of our knowledge, there is no similar solution to the one introduced in this paper for automatically generating topic pages for scientific concepts. Most of the related work falls under the definition extraction task, and this is where we put the focus in this section.
Early work on definition extraction task was focused on rule-based and pattern-matching approaches \cite{hearst1992automatic,klavans2002method,westerhout2009definition}, often resulting in low recall given their limited coverage. Supervised models have also been proposed and shown to be more effective than the rule-based methods for this task \cite{jin2013mining,navigli2010learning,reiplinger2012extracting,roig-mirapeix-etal-2020-definition,kobylinski2008definition}. These models use statistical information regarding concepts, as well as structural information of the sentences such as part of speech (\emph{POS}) tags to distinguish definitional from non-definitional sentences. 
More recent work for definition extraction focused on using neural models for the task \cite{li2016definition,espinosa-anke-schockaert-2018-syntactically,jain2018content,veyseh2020joint,kang2020document,murthy2022accord,veyseh2020does}. Notably, \emph{LSTM} \cite{li2016definition} and a combination of \emph{CNN} and \emph{LSTM} \cite{espinosa-anke-schockaert-2018-syntactically} have been used to learn the structure of definitional sentences. In our work, we also introduce and use a combined \emph{LSTM+CNN} model but, different from \cite{espinosa-anke-schockaert-2018-syntactically}, we capture both semantic (learned from the sentence itself) and structural information within sentences (learned from POS tags). A joint model that encodes sentences and their structure has been used before  in \cite{kang2020document}, but, unlike the task tackled in that work, our definition extraction component assumes that the term is known and tries to detect whether the given candidate sentence is a good definition for the term or not.

\section{Topic Pages Pipeline}
\label{sec:pipeline}

\begin{figure}[t]
\centering
\includegraphics[width=10cm, height=3cm]{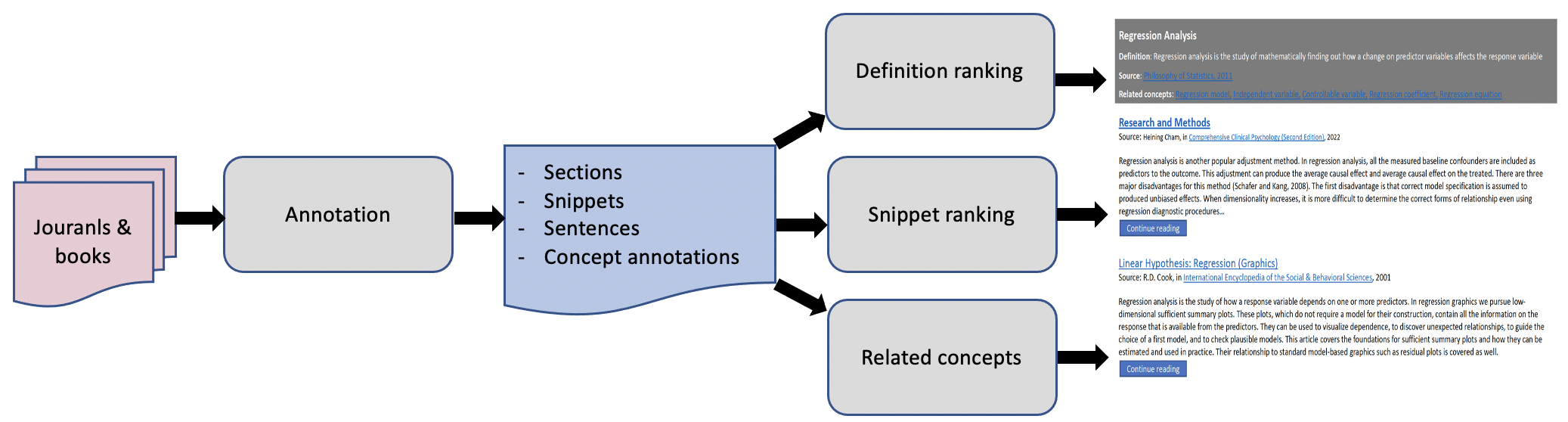}
\footnotesize
\caption{An overview of the topic pages generation pipeline including all essential components. }
\vspace{-10pt}
\label{fig:pipeline}
\end{figure}

There are four main components for generating \emph{Topic Pages}, as shown in Figure \ref{fig:pipeline}: an annotation module, a definition ranking module, a snippet ranking module, and a related concept extraction module.
\vspace{-10pt}

\subsection{Article Annotation }
The annotation module receives content in \emph{XML} format, finds concepts' mentions in articles and books, and then feeds the sentences and snippets mentioning a concept into the subsequent components. Each section in the article is considered a snippet. After we perform sentence splitting, we annotate concepts in sentences by using a simple dictionary look-up against the \emph{Omniscience} taxonomy \cite{malaise2018omniscience} which is a taxonomy of scientific concepts. If an abbreviation for the concept is proposed in the text, such as ``Machine Learning (ML)'', then the abbreviation (ML) is also added as an alias for the concept and is looked in the article. We use the Schwartz and Hearst method \cite{schwartz2002simple} to detect such abbreviations. If multiple concepts partially share some span (of an annotation), we annotate the span with the longest concept and ignore the short annotation. 

\subsection{Definition Ranking and Extraction}
Definitions provide a concise description of the concept. For each concept, and per domain, we rank all the sentences where the concept was annotated and select the top-ranked one as the definition for the concept. We simplify the machine learning task to binary classification where, given a concept and a candidate sentence, the model predicts if it is a good definition for the concept or not. For a target concept, candidate sentences are ranked based on the score the classifier assigns to them and the top-ranked sentence is used as the definition. We use two different models for the definition classification task: an \emph{LSTM+CNN} and a \emph{SciBERT} model. 

\textbf{LSTM+CNN model} Previous work \cite{li2016definition,espinosa-anke-schockaert-2018-syntactically} used \emph{LSTM} \cite{hochreiter1997long} and \emph{CNN} \cite{lecun1998gradient} models to classify sentences in the definition classification task.
We use a combined approach that uses two \emph{LSTMs} and two \emph{CNNs}: one \emph{LSTM} gets the actual sentence as the input and captures the sequential patterns of terms, and the other \emph{LSTM} gets the \emph{POS} tags of the words in the sentence as the input and captures the sequential patterns of syntax in the sentence. One \emph{CNN} gets the actual sentence as the input captures the spatial distribution of terms, and the other \emph{CNN} gets the \emph{POS} tags of the words in the sentence as the input and captures the spatial distribution of grammatical elements. We concatenate the representations learned by each of these models and feed it to a feed-forward \emph{MLP} layer which does the classification, using cross-entropy loss for training.

\textbf{SciBERT model} We use the \emph{SciBERT} model \cite{beltagy-etal-2019-scibert} which is trained on scientific articles. As input, we feed the concept and the candidate sentence separated with a special token ([SEP]) to the \emph{SciBERT} model and get the representation of the [CLS] token. This representation is then fed to a simple feed-forward layer which does the classification, using cross-entropy loss for optimization. 

\subsection{Snippet Ranking}
For a given concept, all snippets annotated with the concept are collected and ranked by a snippet ranking method. The top 10 snippets are used for generating the \emph{Topic Page} for the concept. We use a lexical matching model that scores snippets using a simple location-aware term frequency score as follows:
\begin{equation}
F(c,s) = \frac{tf(c,s)}{|s|} * (1-\frac{l_1(c)}{|s|})    
\end{equation}
where $c$ and $s$ are a concept and a snippet respectively, $tf(c,s)$ is the frequency of $c$ in $s$, $|s|$ is the length of $s$, and $l_1(c)$ is the location of the first occurrence of $c$ in $s$. Hence, the earlier the concept is mentioned in a snippet, the higher the score the snippet would receive.
  
\subsection{Related Concept Extraction}
To find the most relevant concepts to a given concept, we retrieve all co-occurring concepts in snippets. Concepts are then ranked based on the number of their co-occurrence with the target concept and the top $5$ concepts are selected as the related concepts to the target concept.

\section{Results}
\label{sec:results}
%
In this section, we describe the scientific content collection and the used taxonomy (\emph{Omniscience}) that are the basis of the \emph{Topic Pages}. We further discuss the results of the different definition ranking models on two datasets and provide some statistics of the generated \emph{Topic Pages} and usage statistics over time. We leave a large-scale evaluation of the snippet ranking and the related concept extraction modules to future work. 


\subsection{Datasets and baselines} 
We use a collection of articles published in over $2,700$ journals as well as the content of $43,000$ books to generate the \emph{Topic Pages}. This collection contains over $18$ million articles and book chapters in \emph{XML} format. All journals and books belong to different scientific domains. We use the \emph{OmniScience} taxonomy to build the \emph{Topic Pages}, which contain over $700$K concepts for the $20$ domains.

To evaluate the performance of the definition ranking module, we use the \emph{WCL} dataset \cite{navigli2010learning} which contains $4,619$ sentences labeled either as definitional (``good") or non-definitional (``bad") sentences regarding a concept. We follow the same setup as \cite{li2016definition} for training and evaluating models on this dataset.
We additionally use a proprietary dataset containing $43,368$ sentences extracted from articles and books distributed across $8$ different domains and labeled by subject matter experts for the definition evaluation task as either ``good" or ``bad" definitions regarding a concept. We compare the performance of several models including the \emph{LSTM+CNN}, \emph{SciBERT}, Navigli and Velardi \cite{navigli2010learning}, Li et al. \cite{li2016definition}, and Jin et al. \cite{jin2013mining} on the \emph{WCL} dataset. We further evaluate the performance of the best-performing models on the proprietary dataset. For the \emph{LSTM+CNN} model, the batch size is set to $32$, the number of hidden layers of the \emph{LSTM} model is set to $128$, and word embeddings are initiated with \emph{GloVe} and fine-tuned during training. The \emph{MLP} module has a hidden layer with $256$ dimensions trained for $10$ epochs. The \emph{SciBERT} model is trained for $8$ epochs with a batch size of $16$. We perform $10$-fold cross-validation and report the average performance. 

\begin{table}[t]
\centering
\begin{tabular}{lccc}
\hline
\textbf{Model} & \textbf{Precision} & \textbf{Recall} & \textbf{F1}\\
\hline
Jin et al. \cite{jin2013mining} & 0.92 & 0.79 & 0.85\\
Li et al. \cite{li2016definition} & 0.90 & 0.92 & 0.91\\
Navigli and Velardi \cite{navigli2010learning} & \textbf{0.99} & 0.61 & 0.85\\ 
LSTM+CNN & 0.94 & 0.91 & 0.92\\ 
SciBERT & 0.94 & \textbf{0.93} & \textbf{0.93}\\
\hline
\end{tabular}
\caption{Performance of different definition classification models on the WCL dataset in terms of macro-averaged precision, recall, and F1.}
\vspace{-10pt}
\label{table:def-general-results}
\end{table}

\subsection{Results of the definition extraction models} 
Table \ref{table:def-general-results} shows the performance of different models on the \emph{WCL} dataset. This dataset is extracted from Wikipedia and most of the Wikipedia-based definitions follow a similar structure, making them easy to classify. The \emph{SciBERT} model achieves the best \emph{F1} score on this dataset. Navigli and Velardi \cite{navigli2010learning} have higher precision than all models but a very low recall compared to \emph{SciBERT}. The higher performance of the \emph{SciBERT} model compared to the \emph{LSTM+CNN} model shows that \emph{SciBERT} can learn both sequential and spatial distribution of words in definitional sentences as well as the structural information within such sentences.

We further evaluate the performance of the top-performing models (\emph{SciBERT} and \emph{LSTM+CNN}) on the proprietary dataset which is much larger than the \emph{WCL} dataset; results are shown in Table \ref{table:def-domain-results}. This dataset contains definitions from various sources. Unlike Wikipedia-based definitions, definitions extracted from different books and journals do not follow a similar structure which makes the classification task more difficult, hence the lower performance of the two models compared to the \emph{WCL} dataset. The \emph{SciBERT} model outperforms the \emph{LSTM+CNN} model on this dataset as well across all domains. This again confirms the ability of the \emph{SciBERT} model in modeling semantics and the structure of definitions. 
Moreover, \emph{SciBERT} has consistently higher performance than the \emph{LSTM+CNN} on all individual domains except \emph{Social Sciences}. As \emph{SciBERT} is pre-trained on publications in the biomedical and computer science domains the low performance of this model on domains such as \emph{Social Sciences} may be attributed to this fact. On the other hand, as the results show, \emph{SciBERT} performs better on domains such as \emph{Chemistry} and \emph{Material Sciences} as such domains are closer to its trained domains.

\begin{table}
\centering
\footnotesize
\begin{tabular}{lll}
\hline
\textbf{Concept} & \textbf{Definition} & \textbf{Error source}\\
\hline
Association List & An association list is simply a list of name value pairs. &Too generic\\
Hierarchical DB & In a hierarchical DB relationships are & Too generic \\
& defined by storage structure&\\
Habilitation & The acquisition of abilities not possessed previously. & Too specific\\
Sample Space& the set of all possible outcomes in a probability model&Partially good\\
\hline
\end{tabular}
\caption{Example of errors (false positives) of the SciBERT-based models.}
\label{table:examples}
\end{table}

Other than the domain difference, the additional errors should be attributed to the inherent difficulty of the task. 
Based on our analysis, the biggest sources of errors are the false positives which are mainly caused by generic, specific, or partially good definitions.
Table \ref{table:examples} shows examples of definitions wrongly labeled by the \emph{SciBERT} model and the possible explanation for the errors.
Generic definitions are good definitions but they cover a very broad aspect of the concept.
Specific definitions are also good definitions but they contain unnecessary additional information.
Partially good definitions cover only some essential aspects of the concept.
All these cases are labeled as ``bad definitions" by the subject-matter experts but detected as ``good definitions" by the model.
To handle such cases, the model should have an understanding of the generality or specificity of the concept which can be quite challenging to model.

\begin{table}[!t]
    \centering
    \footnotesize
    \begin{tabular}{lccccccc}
        \hline
        \multirow{2}{*}{\textbf{domain}} & \multicolumn{3}{c}{\textbf{SciBERT}} & & \multicolumn{3}{c}{\textbf{LSTM+CNN}}\\
        \cmidrule{2-4}\cmidrule{6-8}
         & \textbf{Precision} & \textbf{Recall}  & \textbf{F1} & & \textbf{Precision} & \textbf{Recall}  & \textbf{F1} \\
        \hline 
        Chemistry & 0.78 & 0.80 & 0.79 & & 0.69 & 0.68 & 0.68 \\
        Earth Sciences & 0.80 & 0.84 & 0.82 & & 0.66 & 0.64 & 0.65 \\
        Material Sciences & 0.80 & 0.88 & 0.83 & & 0.50 & 0.49 & 0.49 \\
        Computer Science & 0.56 & 0.60 & 0.58 & & 0.43 & 0.48 & 0.45 \\
        Social Sciences & 0.39 & 0.43 & 0.41 & & 0.38 & 0.46 & 0.42 \\
        \hline
        All domains & \textbf{0.79} & \textbf{0.78} & \textbf{0.78} && 0.70 & 0.69 & 0.69   \\
        \hline
    \end{tabular}
    \caption{Performance of the \emph{LSTM+CNN} and \emph{SciBERT} models on five domains.}
    \vspace{-10pt}
\label{table:def-domain-results}
\end{table}


The \emph{Topic Pages} product contains over $363,000$ topic pages in $20$ different scientific domains. 
Topic pages have over $23$ million visits per month making them one of the popular knowledge bases among researchers and students.
There are about $63,000$ concepts without a definition on \emph{Topic Pages} mostly due to the bad performance of the current production model (\emph{LSTM+CNN}) in some domains.



\section{Conclusions and Discussion}
\label{sec:limits}
\vspace{-10pt}
In this paper, we introduced \emph{Topic Pages}, a publicly available knowledge base for scientific concepts with their definitions, most relevant concepts, and snippets providing more context around them. 
We described all the major components combined to build this resource.
The pipeline for generating \emph{Topic Pages} can be used on top of any document collection as well as a taxonomy to build a similar resource in any domain.
With over $363,000$ topic pages in $20$ different scientific domains, and more than $23$ million unique visitors per month, \emph{Topic Pages} are one of the popular knowledge bases among researchers and students. We described all major components of the pipeline for extracting different pieces of information necessary to generate the pages. In this work, we mainly focused on building a high-performance definition extraction model. To this end, we used an \emph{LSTM+CNN} and a \emph{SciBERT} model. Empirical evaluation shows that both models can outperform existing models for the definition classification and extraction task. 
However, the \emph{SciBERT} model still needs to be improved for domains such as \emph{Social Sciences}. The biggest drawback of using \emph{SciBERT} for such domains is that this model is pre-trained on mostly biomedical articles and, therefore, it cannot model all other domains as well. As a future work, we would like to exploit the concepts and their definitions extracted from Wikipedia as well as expand our dataset to further fine-tune the \emph{SciBERT} model for such domains.
As another future work, we are going to use the click-through data we have collected as a proxy to train supervised models for related concept extraction and snippet ranking components.

\newpage
%
 \bibliographystyle{splncs04}
 \bibliography{bibliography}

\end{document}